%% file: paper.tex
\documentclass[letterpaper,10pt,conference]{ieeeconf}

\IEEEoverridecommandlockouts
\usepackage{amsmath,amssymb}
\usepackage{cite}
\usepackage{graphicx}
\usepackage{microtype}
\usepackage{url}
\usepackage{balance}
\usepackage{booktabs,array}
\usepackage{algorithm}
\usepackage{algpseudocode}
\usepackage{dblfloatfix}
\usepackage{xcolor}
\usepackage{eso-pic}

\title{\LARGE \bf
C2Flight: A Closed-Loop Co-Simulation Framework for Early-Stage Evaluation of Heterogeneous UAV Computing Systems
}

\author{Xizhu Chen, Shuyuan Zhang, Yixuan Li, Huazhong Yang, Yu Wang, Jincheng Yu, Hongyang Jia}

\begin{document}

\AddToShipoutPictureFG*{%
    \AtPageUpperLeft{%
        \put(36,-22){%
            \parbox[t]{\dimexpr\paperwidth-72pt\relax}{\centering\footnotesize
                This work has been submitted to the IEEE for possible publication.
                Copyright may be transferred without notice, after which this version may no longer be accessible.
            }%
        }%
    }%
}

\maketitle
\thispagestyle{empty}
\pagestyle{empty}

\input{Sections/0_Abstract}
\input{Sections/1_Introduction}

\input{Sections/2_Related_Work}
\input{Sections/3_C2Flight_Framework}
\input{Sections/4_Evaluation}
\input{Sections/5_Case_Study}
\input{Sections/6_Conclusion}

\balance
\bibliographystyle{IEEEtran}
\bibliography{cite}

\end{document}

%% file: Sections/0_Abstract.tex
\begin{abstract}

Early-stage design of heterogeneous UAV computing systems requires evaluating how algorithm, hardware, communication,and system topology affect closed-loop flight before the complete system is available.
However, design choices interact through the robotic feedback loop, making task performance difficult to predict from isolated component metrics, while existing execution-driven approaches postpone system-level design feedback until the computing system is integrated.
This work presents C2Flight, a closed-loop co-simulation framework that represents target systems as parameterized dataflow graphs and integrates models of hardware under design with independently accessible physical hardware through a unified node interface.
Its graph runtime captures the combined effects of buffering, activation, resource sharing, communication, and local timing on closed-loop flight behavior. Validation against onboard and wireless offboard physical computing pipelines confirms its timing fidelity and closed-loop prediction accuracy. Two case studies demonstrate early-stage exploration of deployment and accelerator bit-width choices.
These results show that C2Flight exposes system-level design trade-offs before full-system integration, enabling rapid early-stage design refinement.

\end{abstract}


%% file: Sections/1_Introduction.tex
\section{Introduction}

{\looseness=-1
Enabling autonomous unmanned aerial vehicles (UAVs) to perform perception and planning under stringent payload, power, and timing constraints increasingly demands heterogeneous computing systems~\cite{geles2024agile,raju2025edgeflownet,scarciglia2025mapfree,kapoor2025visafe}.
Designing these systems entails tightly coupled decisions over algorithm selection, hardware architecture, compute placement, and system topology, yielding a vast cross-layer design space~\cite{paredesvalles2024neuromorphic,chen2024fogros2ft, bakhshalipour2024tartan, huang2025daducorki}.
Therefore, exploring this design space through full-system implementation would be prohibitively slow and costly, making early-stage system-level evaluation essential to close the design feedback loop while key decisions remain reversible. 
\par}

However, evaluating the closed-loop task performance of autonomous UAVs is particularly challenging during the early-stage design of their heterogeneous computing systems.
The effects of algorithm, hardware, and communication choices propagate through the robotic feedback loop, creating feedback-dependent trade-offs that make task performance difficult to infer from isolated component metrics~\cite{krishnan2022autopilot,gog2021pylot,ali2025realtime}. 
For example, low-bit quantization can reduce computation and communication overhead but degrade perception accuracy, thereby obscuring its net impact on task performance. 
Effective early-stage evaluation must therefore capture this closed-loop propagation, even while target algorithms and hardware are still under development and communication links remain unavailable.

Existing approaches capture these coupled effects in a closed loop yet provide limited support for early-stage evaluation~\cite{gog2021pylot, srivastava2025vizflyt, boroujerdian2018mavbench}. Pylot~\cite{gog2021pylot} performs execution-driven evaluation of an end-to-end autonomy pipeline, VizFlyt~\cite{srivastava2025vizflyt} emphasizes hardware-in-the-loop (HIL) testing with a physical aerial robot, and MAVBench~\cite{boroujerdian2018mavbench} combines executed workloads on embedded hardware with closed-loop vehicle simulation. 
Nevertheless, they all rely on complete integration of computing systems, delaying design feedback until substantial development effort has been invested. This leaves early design exploration open-loop at the system level, with components optimized locally without insight into their combined impact on robotic performance.

\begin{figure}[t]
    \centering
    \includegraphics[width=\linewidth,trim=3pt 3pt 3pt 3pt,clip]{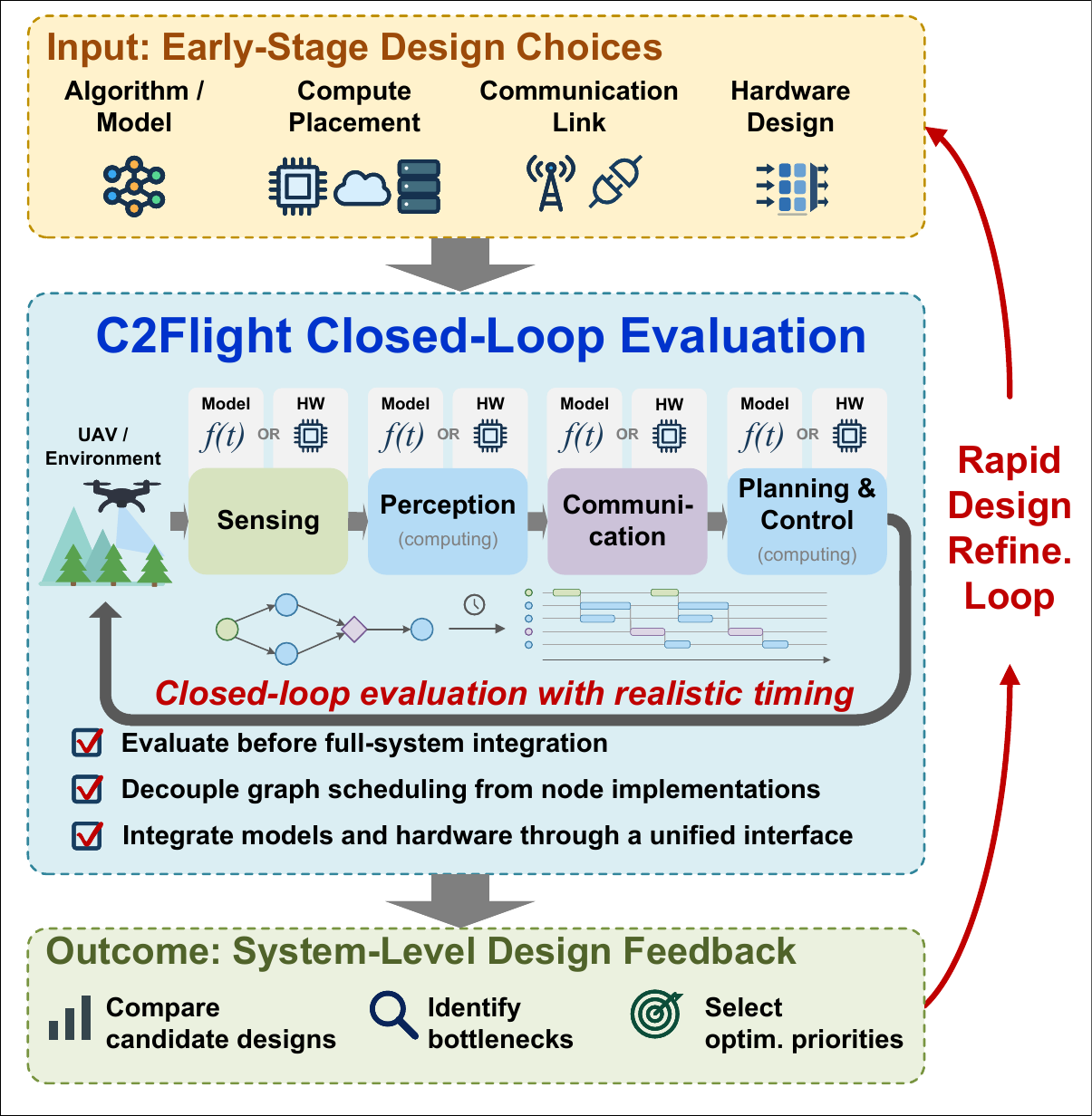}
    \caption{
        \textbf{Early-Stage Design of UAV Computing Systems with C2Flight.} Candidate algorithms, compute placement, communication links, and hardware configurations are represented by a parameterized graph, while a unified interface combines component models with available hardware for closed-loop evaluation. The resulting system-level feedback supports candidate comparison, bottleneck identification, and optimization prioritization, closing the design feedback loop before full-system integration.
    }
    \label{fig:introduction}
\end{figure}

{\looseness=-1
To overcome these challenges, we present C2Flight, a closed-loop co-simulation framework for early-stage evaluation of heterogeneous UAV computing systems. By combining available implementations with models of unfinished components and links, C2Flight evaluates their coupled effects within the robotic feedback loop before full integration. The resulting design feedback helps compare candidate designs, identify bottlenecks, and prioritize optimization, as shown in Fig.~\ref{fig:introduction}.
\par}

\newpage
The main contributions of this work are as follows:
\begin{itemize}
    \item We design a parameterized dataflow graph defined by computation and interaction boundaries, allowing components to be replaced without rebuilding the entire system while preserving timing dependencies for accurate end-to-end latency modeling. 
    \item We propose a unified interface that integrates common timing profiles and hardware interaction protocols, enabling C2Flight to evaluate computing systems with hybrid configurations composed of modeled and real-hardware nodes.
    \item We validate C2Flight's timing fidelity against a physical sensing-to-command pipeline and conduct two case studies, revealing how accuracy and control latency jointly shape system-level design trade-offs.
\end{itemize}

%% file: Sections/2_Related_Work.tex
\section{Related Work}

{\looseness=-1
Existing closed-loop UAV evaluation platforms fall into two categories: simulation-based platforms and HIL systems. Simulation platforms such as AirSim~\cite{shah2018airsim}, Flightmare~\cite{song2020flightmare}, and CrazySim~\cite{llanes2024crazysim} provide rendered environments, sensor and dynamics models, and flight-control interfaces for algorithm development. 
HIL systems execute algorithms on physical computing platforms while simulating sensor inputs, vehicle dynamics, or both. Some also incorporate a physical flying vehicle. 
MAVBench~\cite{boroujerdian2018mavbench} couples ROS-based autonomy workloads on embedded platforms with AirSim and uses energy and battery models to relate computing behavior to mission time and energy. 
Agilicious~\cite{foehn2022agilicious} combines real quadrotor flight with photorealistic sensor simulation, retaining physical dynamics and proprioceptive sensing while allowing visual conditions to vary independently. 
FlightForge~\cite{capek2025flightforge} generates large-scale procedural environments and returns observations rendered from a real vehicle's pose to its onboard navigation and control stack. 
By deriving computing behavior from actual pipeline execution, these systems capture software, hardware, and communication effects in a closed loop, but require the corresponding subsystems to be implemented and integrated.
\par}


Beyond characterizing existing UAV computing platforms, prior early-stage co-design frameworks extend closed-loop simulation from algorithms to candidate computing architectures before fabrication. 
AutoPilot~\cite{krishnan2022autopilot} combines Bayesian optimization with policy and accelerator simulation to jointly explore autonomy algorithms and hardware configurations under sensor, payload, and mission-energy constraints.
RoS\'E~\cite{nikiforov2023rose} couples AirSim with FPGA-accelerated register-transfer-level (RTL) simulation in FireSim to assess candidate SoCs with the complete robotics stack before fabrication. It synchronizes time and data transfer across the simulators while exposing computation, data movement, and resource contention through the candidate RTL designs. 
EAGERx~\cite{heijden2025eagerx} represents abstract objects as engine-specific subgraphs for simulation or real hardware and coordinates them through a decentralized message-count protocol, although this synchronization assumes fixed node rates and deterministic delays within each episode. 
These frameworks adopt diverse system representations to support early hardware exploration and model--hardware integration, yet remain limited in composing partially implemented heterogeneous systems and capturing execution-dependent computation and communication effects within a unified closed loop.


Extending closed-loop analysis to partially implemented heterogeneous UAV computing systems requires composable representations of computation, communication, and their runtime interactions. C2Flight provides this capability through a graph-based system abstraction and a unified model--hardware interface. The graph decouples system structure from node implementations, while the interface maintains consistent interaction and timing semantics across modeled and physical nodes, allowing components to transition from models to hardware without rebuilding the closed-loop system.

%% file: Sections/3_C2Flight_Framework.tex
\section{C2Flight Framework}
\label{sec:framework}

\begin{figure*}[t]
    \centering
    \includegraphics[
        width=0.92\textwidth,
        trim=3pt 3pt 3pt 3pt,
        clip
    ]{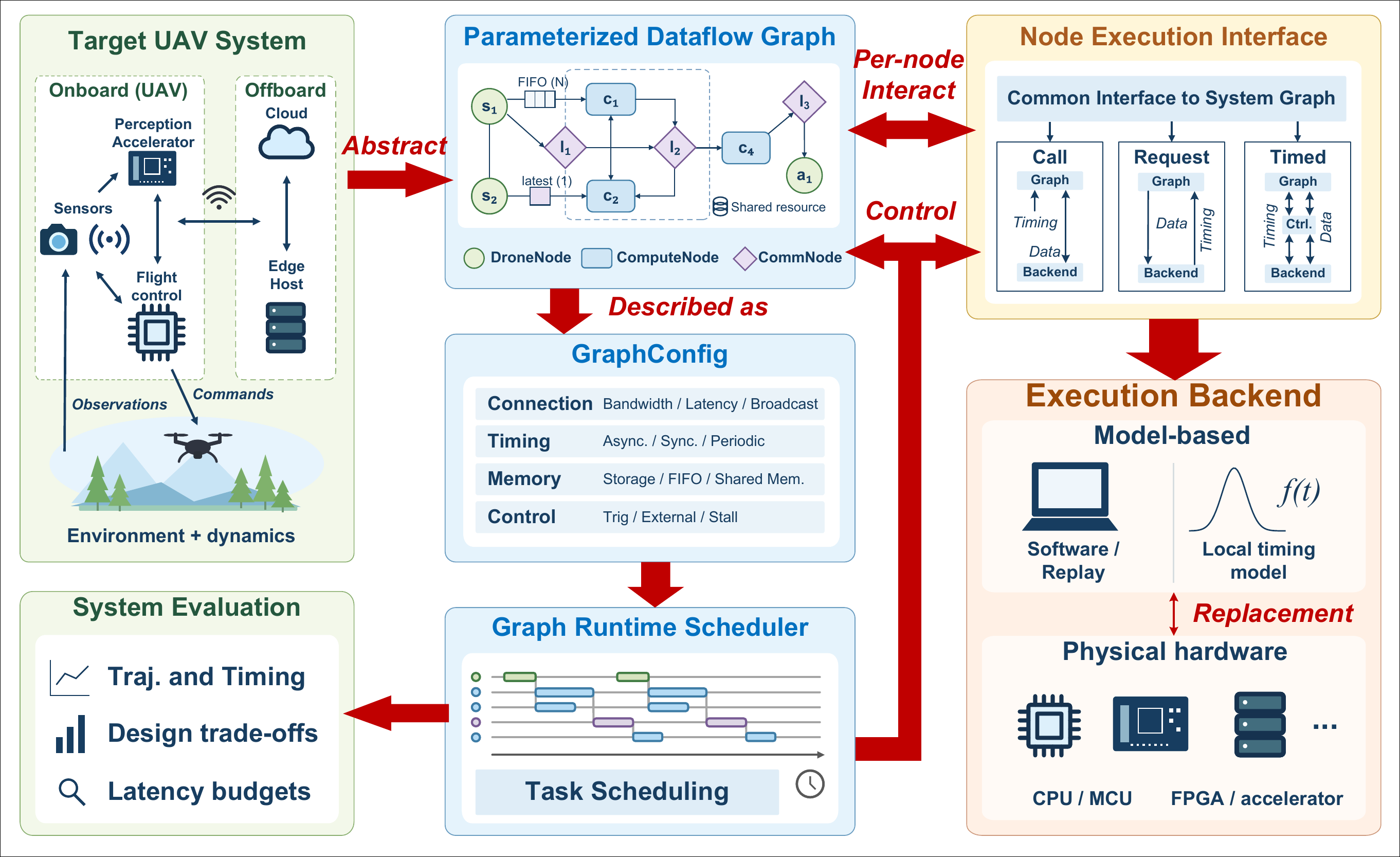}
    \caption{
        Overview of C2Flight.
        The graph exposes computation and interaction boundaries.
        \texttt{GraphConfig} defines inter-node policies, while \texttt{NodeConfig} selects implementations and local timing.
        The unified interface connects these descriptions to \texttt{GraphRuntime} and the environment adapter.
        Experimental access overhead is recorded separately and excluded from target latency where the execution contract permits this separation.
        Compatible replacements preserve declared port semantics and graph policies for system-level comparison.
        Solid blue arrows indicate data exchange; dashed blue arrows indicate runtime coordination and backend binding.
    }
    \label{fig:framework}
\end{figure*}

Fig.~\ref{fig:framework} presents an overview of C2Flight, a closed-loop co-simulation framework designed for accurate early-stage evaluation of heterogeneous UAV computing systems. C2Flight takes as input a description of the target computing system together with a simulated flight environment. The system description specifies its topology, inter-node policies, and each component's implementation and timing source, while the environment generates sensor observations and updates the vehicle state in response to control commands. Each run produces timing traces and flight trajectories to analyze end-to-end latency, identify bottlenecks, and compare task-level performance across candidate systems.

{\looseness=-1
At its core, C2Flight represents a target UAV computing system as a parameterized graph, decoupling computation and interaction boundaries from current node implementations. The graph comprises \texttt{ComputeNode}s, \texttt{CommNode}s, and a \texttt{DroneNode}. Two orthogonal configurations specify inter-node interactions and node implementations separately. \texttt{GraphConfig} defines how these nodes exchange data, activate one another, and share resources, while \texttt{NodeConfig} selects each node's functional backend, timing source, and access method. A unified node interface exposes consistent functional and timing events for modeled and real-hardware nodes, enabling C2Flight to co-simulate both within the same UAV workload.
\par}

During a closed-loop run, \texttt{GraphRuntime} instantiates the configured graph and schedules its events on a shared target-time axis. Observations generated by the environment enter through the \texttt{DroneNode} and propagate through computation and communication nodes. Intermediate results become available according to their execution and transfer timing, and the resulting commands return to the flight dynamics to update the vehicle state and generate subsequent observations. Because the graph structure and its interaction policies remain unchanged when node backends are replaced, available software and hardware can be combined with models of unfinished components and links without rebuilding the surrounding system. The remainder of this section details the mechanisms underlying this workflow.

\subsection{System Abstraction}
\label{sec:system_abstraction}

{\looseness=-1
C2Flight represents a target system as $G=(V, E;\Theta_g)$, where $V$ contains logical nodes, $E$ defines directed connections between their ports, and $\Theta_g$ specifies inter-node interaction rules.
Nodes are defined by computation and interaction boundaries rather than physical device boundaries, allowing different stages within the same device to be modeled separately.
\par}

{\looseness=-1
Nodes are classified into three types. \texttt{ComputeNode} represents processing, \texttt{CommNode} exposes independently timed communication, and \texttt{DroneNode} connects sensing and control to the environment.
For example, preprocessing and inference on one platform can be separate compute nodes connected through an explicit communication node.
Several nodes may share a backend while retaining their identities; shared target resources must still be declared explicitly.
\texttt{GraphConfig} defines interactions, whereas \texttt{NodeConfig} selects functionality, local timing, and experimental access without redefining those policies.
Thus, modeled links connect independently available components, while compatible replacements retain the surrounding graph rather than requiring a complete system rebuild.
With this abstraction, a heterogeneous UAV computing system can be partitioned into nodes, connected according to their interactions, and configured using predefined options to construct a parameterized dataflow graph for closed-loop simulation.
\par}

\subsection{Parameterized Dataflow}
\label{sec:parameterized_dataflow}

After system abstraction, dataflow between nodes is expressed through the graph configuration $\Theta_g=(\Theta_d,\Theta_a,\Theta_r;\Theta_t)$, which combines three interaction groups with common time conventions.
The parameter choices in Table~\ref{tab:graph_parameters} expand into common rules for data transfer and node execution, without requiring user-written, system-specific scheduling functions.

\begin{table*}[t]
    \centering
    \caption{Representative choices for the three graph interaction groups and their shared time convention. Node implementations, local timing sources, and access settings belong to node configurations.}
    \label{tab:graph_parameters}
    \footnotesize
    \setlength{\tabcolsep}{4pt}
    \renewcommand{\arraystretch}{1.12}
    \begin{tabular}{
        @{}
        >{\raggedright\arraybackslash}p{0.24\textwidth}
        >{\raggedright\arraybackslash}p{0.24\textwidth}
        >{\raggedright\arraybackslash}p{0.47\textwidth}
        @{}
    }
        \toprule
        \textbf{Group} & \textbf{Parameter} & \textbf{Representative choices} \\
        \midrule

        \textbf{Data exchange} $(\Theta_d)$ & Port storage mode & Queued storage, latest-value storage, direct handoff \\
        & Full-buffer policy & Blocking, dropping new inputs, dropping oldest inputs \\
        & Read policy & Data consumption, data retention \\
        & Packet selection policy & FIFO order, latest arrival, latest source timestamp \\
        & Connection mode & Direct transfer, broadcast, tagged routing \\

        \midrule
        \textbf{Node activation} $(\Theta_a)$ & Trigger type & Data-driven, periodic, completion-triggered, external \\
        & Pending-trigger policy & Queueing, keeping the latest, skipping while busy \\
        & Input grouping & Key matching, timestamp matching, snapshots, batches \\

        \midrule
        \textbf{Execution and resources} $(\Theta_r)$ & Execution pattern & Task execution, pipelined execution, chunk streaming \\
        & Resource type & Capacity slots, shared service rate \\
        & Slot arbitration policy & FIFO, priority, round-robin, deadline \\

        \midrule
        \textbf{Time convention} $(\Theta_t)$ & Same-time event order & Publication first, trigger first \\

        \bottomrule
    \end{tabular}
\end{table*}

\emph{Data exchange} $(\Theta_d)$ defines port storage, full-buffer handling, read and packet selection policies, and connection modes. Packets retain input identities and source timestamps, while $t^{\mathrm{available}}$ marks their earliest eligibility for selection. Retention permits repeated reads without refreshing source time; blocking makes delivery depend on downstream capacity. \emph{Node activation} $(\Theta_a)$ defines trigger types, pending-trigger policies, and input grouping. These policies determine when execution is requested and which observations it uses, independently of the processing algorithm. Pending-trigger handling further determines whether requests accumulate while a node is busy. \emph{Execution and resources} $(\Theta_r)$ specifies execution patterns, resource types, and slot arbitration. Pipelining separates initiation from completion, so task latency alone does not determine throughput. For shared-rate resources, completion depends on remaining work and allocated service. \emph{Time convention} $(\Theta_t)$ selects publication-first or trigger-first priority for otherwise unordered events within causal rounds. This choice can change which data a coincident trigger selects.

Together, these configuration groups express the system's dataflow as executable rules, while node functionality and local timing sources remain independently configurable. The following subsection describes how \texttt{GraphRuntime} schedules the resulting parameterized dataflow graph within the UAV feedback loop.

\subsection{Closed-Loop Scheduling}
\label{sec:closed_loop_scheduling}

\begin{algorithm}[t]
    \caption{Execution of the configured graph by \texttt{GraphRuntime}}
    \label{alg:closed_loop}
    \footnotesize
    \begin{algorithmic}[1]
        \Require $G$, node configurations, controlled backends
                 $\mathcal{B}_t$, environment, horizon $T$
        \State Initialize graph state, node bindings, environment,
               and event queue $Q$
        \State $t \gets 0$
        \Loop
            \Repeat
                \State Process the next due causal round using $\Theta_t$
                \State Apply $\Theta_d$ transfers and $\Theta_a$ activation
                       and input selection
                \State Admit tasks under $\Theta_r$; execute through
                       the node interface
                \State Update downstream readiness; collect node events
                \State Check errors and the same-time iteration limit
            \Until{stable at $t$, or a stopping condition is reached}

            \If{failure, environment termination, or $t \geq T$}
                \State Report the corresponding status; \textbf{break}
            \EndIf
            \If{no future graph or backend event}
                \State Report completion or unresolved waits; \textbf{break}
            \EndIf

            \State $t_Q \gets$ next graph-event time
            \State $t_B \gets$ earliest next declared target-clock boundary
            \State $h \gets \min\{t_Q,t_B,T\}$
            \ForAll{$b\in\mathcal{B}_t$}
                \State $(r_b,s_b,\mathcal{E}_b)
                       \gets b.\texttt{advance}(h)$
                \State Require $r_b=s_b=h$
                \State Reject skipped interior events;
                       validate and enqueue $\mathcal{E}_b$
            \EndFor
            \State Advance the environment state to $h$; $t\gets h$
        \EndLoop
        \Statex \textit{Due events include sensing and command submission.}
        \Statex \textit{Absent event/clock times are $+\infty$.}
        \Statex \textit{A violated backend contract terminates the run.}
    \end{algorithmic}
\end{algorithm}

{\looseness=-1
\texttt{GraphRuntime} constructs state from $\Theta_g$ and combines node events as summarized in Algorithm~\ref{alg:closed_loop}.
Pending native admissions reserve acceptance/start resources until acquisition; resources configured for output are not pre-reserved.
Within each causal round, $\Theta_t$ orders otherwise unordered publication/trigger events and uses stable tie-breaking; newly generated same-time events enter subsequent rounds.
Blocked states are retained, and unresolved waits or nonconvergent zero-time activity terminate with diagnostic status rather than artificial delay.
\par}

For assigned-offset tasks, $t_{k,p}^{\mathrm{ready}}=t_k^{\mathrm{start}}+\Delta_{k,p}^{\mathrm{ready}}$ marks local output readiness.
Receiver visibility begins only after a valid handoff, and selection still follows the configured activation and input policies. Controlled backends advance to a common boundary limited by the next graph event, declared target-clock edge, and horizon.
Declared clock boundaries must cover every possible externally visible interaction, rather than being arbitrary polling intervals.
Each backend must reach and confirm that common boundary; skipped interior events are rejected, without arbitrary early-return coordination or rollback.
Timestamp precision does not determine advancement granularity.

{\looseness=-1
In the evaluated environment adapter, Crazyflow advances vehicle dynamics and low-level control, while AirSim renders observations at supplied poses in ComputerVision mode.
Rendering and RPC wall time do not advance the modeled state; renderer-history effects require separate treatment.
Setpoint readiness, interface submission, and controller consumption are distinct events.
Graph policies determine submission, while controller execution and plant integration follow environment schedules without an implicit extra command-update period.
\par}

\subsection{Node Execution Interface}
\label{sec:node_execution_interfaces}

\begin{figure}[t]
    \centering
    \includegraphics[
        width=\columnwidth,
        trim=2pt 2pt 2pt 2pt,
        clip
    ]{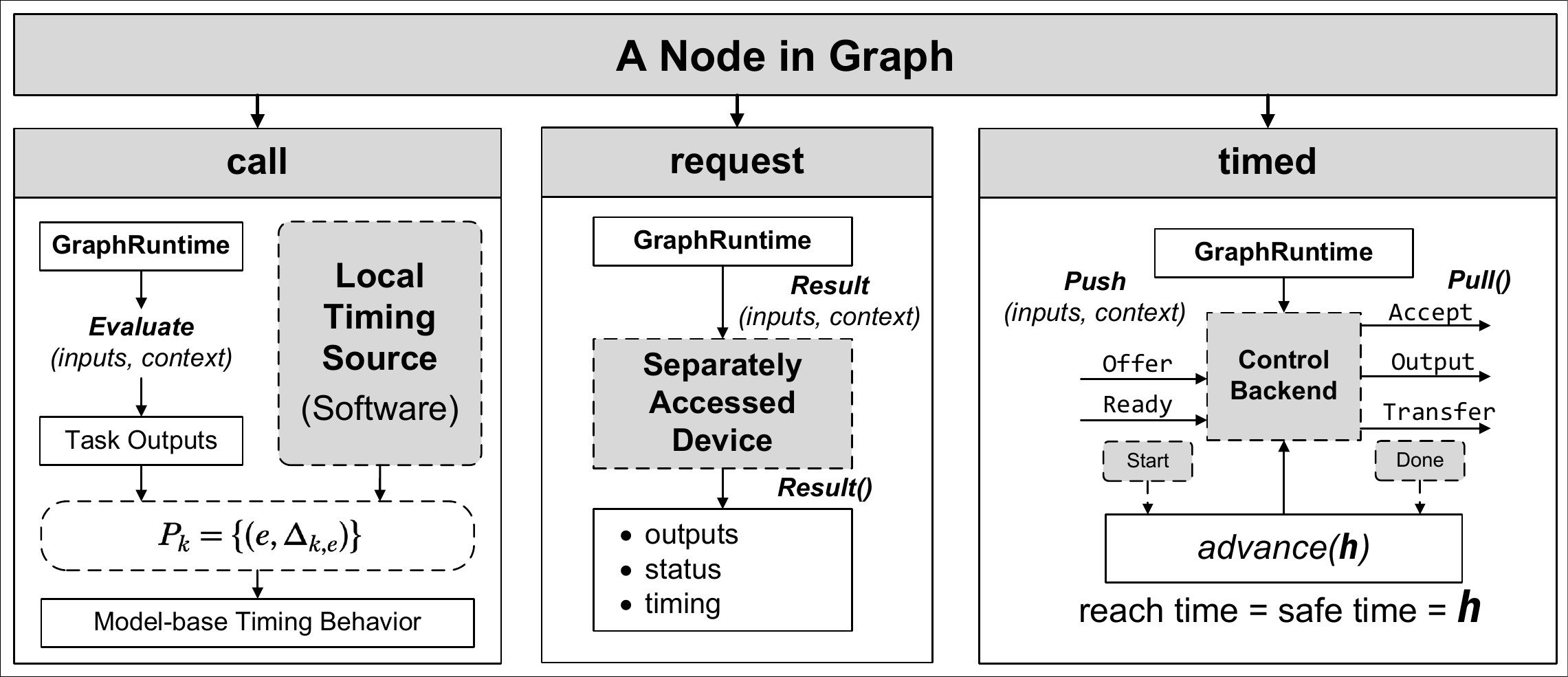}
    \caption{
        Three backend classes under the unified node execution interface.
        (a) \texttt{call} returns locally computed outputs.
        (b) \texttt{request} obtains device outputs and optional measurements.
        For both classes, the runtime combines local timing with graph policies to determine target events.
        (c) \texttt{timed} reports controlled boundary events, receives task-level transfer authorizations, and confirms common-boundary progress.
        The classes share port-event meanings but differ in timing and state-control responsibilities.
    }
    \label{fig:node_interfaces}
\end{figure}

The node execution interface defines common task identities, port formats, and boundary events for modeled and physical backends.
\texttt{NodeConfig} selects a backend class---\texttt{call}, \texttt{request}, or \texttt{timed}---and its implementation, local timing, and access method (Fig.~\ref{fig:node_interfaces}).

{\looseness=-3
\texttt{call} covers local software, replay, and forwarding.
\texttt{evaluate(inputs, context)} processes the selected inputs and returns functional outputs, while fixed values, distributions, traces, or analytical models supply local timing.
For assigned event times, profiles provide offsets from task start.
\texttt{GraphRuntime} combines these with graph policies, without duplicating graph-owned waiting.

\texttt{request} covers task-based hardware access through identified \texttt{CALL}--\texttt{RESULT} exchanges.
Outputs use configured profiles or local measurements with validated boundaries and clock-unit conversion.
Adapters must preserve logical state order and establish that excluding access time leaves target behavior unchanged.
Target concurrency remains graph-scheduled rather than requiring concurrent device calls.

\texttt{timed} covers controlled backends executing internal queues, pipelines, or contention.
Through \texttt{push}/\texttt{pull}, the graph issues \texttt{OFFER}, and the backend reports \texttt{ACCEPT}, \texttt{OUTPUT}, and \texttt{TRANSFER} for input acceptance, output readiness, and completed handoff.
After \texttt{OUTPUT}, the graph grants task-level \texttt{READY} when downstream conditions permit transfer.
The backend must retain outputs until transfer and preserve controlled state during host pauses.
At each common boundary $h$, \texttt{advance(h)} confirms that the reached time and safe time are equal to $h$; earlier reports are rejected, while same-time interactions remain allowed.

The abstraction rests on boundary behavior: data and source identities, event times, resource lifecycles, and subsequent state-dependent responses.
With initial conditions, external inputs, environment, event ordering, and random realizations fixed, replacements matching these responses for every admissible input and acceptance history preserve observable transfers and control inputs.
Each interaction has one execution owner, avoiding duplicate simulation.
These shared boundaries support compatible component replacement without rebuilding the surrounding graph.
\par}

%% file: Sections/4_Evaluation.tex
\section{Evaluation}
\label{sec:evaluation}

We validate C2Flight's timing fidelity and closed-loop prediction accuracy with hybrid configurations of modeled and real-hardware nodes, using integrated physical sensing-to-command pipelines in onboard and offboard UAV computing systems as references.

\subsection{Experimental Setup}
\label{sec:eval_setup}

We build onboard and offboard computing systems on the Crazyflie platform with an AI-deck, using AirSim~\cite{shah2018airsim} for rendering and Crazyflow~\cite{schuck2026crazyflow} for dynamics.
The \textbf{onboard system} follows the PULP-DroNet~\cite{palossi2019pulpdronet} deployment: the Himax camera captures images, GAP8~\cite{flamand2018gap8} runs quantized DroNet~\cite{loquercio2018dronet} for perception, UART transfers the inference results, and STM32 performs navigation and generates setpoints.
These four stages are mapped to separate graph nodes within C2Flight, with graph-level buffering policies governing STM32 input selection and triggering rules capturing its conditional waiting.
In the \textbf{offboard system}, a workstation equipped with an NVIDIA GeForce RTX 4070 GPU runs a ROS-based pipeline that receives Himax images over Wi-Fi, runs PULP-DroNet at FP32 precision, performs navigation, and sends commands through Crazyradio to the STM32 for setpoint generation.
The pipeline comprises three asynchronously executing ROS processes for image reception, computation, and command transmission.
The three ROS processes and STM32 command handling are mapped to compute nodes, whereas Wi-Fi, ROS messaging, and Crazyradio are represented as communication nodes. Graph-level resource constraints capture the Wi-Fi transmission-buffer capacity, while the host perception node's input buffer retains only the latest pending image.

\subsection{Evaluation Configurations}
\label{sec:eval_configs}
{\looseness=-1
\textbf{GT (Ground Truth)} is obtained from fully integrated physical sensing, computing, and communication chains, with end-to-end timing measured from sensor input to command generation.
For closed-loop evaluation, the chain processes rendered observations, and each command's measured age determines when it is injected into the simulation.
We evaluate two C2Flight configurations against this reference.
\textbf{CH (C2Flight Hybrid)} connects independently accessed computing backends, both modeled and real-hardware.
The onboard configuration uses GAP8 and STM32, whereas the offboard configuration uses the RTX 4070 workstation and STM32.
This configuration enables independent hardware platforms to be rapidly combined for closed-loop evaluation and profiling, even before their performance is fully characterized.
\textbf{CM (C2Flight Model)} retains the same target graph and runs equivalent software functions, using independent component timing profiles rather than per-execution hardware measurements.
This configuration supports closed-loop evaluation of hardware designs under development, allowing designers to explore the design space and identify the best-performing configurations.
\textbf{SM (Sequential Model)} uses the same functions and component timing models as CM but replaces graph-based scheduling with serial delay accumulation, solely as an ablation in the timing-fidelity evaluation.
\par}

The closed-loop experiments additionally include two comparison configurations to assess the value of system-level timing and interaction modeling.
\textbf{SC (Subsystem-Centric)} represents early-stage closed-loop evaluation centered on the computing subsystem, as exemplified by RoS\'E~\cite{nikiforov2023rose}.
Our SC configuration retains GAP8 or host execution and basic interface costs while using simplified timing models for sensing and peripheral task interactions.
\textbf{SW (Software-Only)} represents only conventional software validation, combining algorithmic behavior and vehicle dynamics without modeling target computation or communication delays.

\begin{table}[t]
    \centering
    \caption{Command-age statistics and prediction errors}
    \label{tab:eval_timing}
    \label{tab:eval_a2_timing}
    \footnotesize
    \setlength{\tabcolsep}{1pt}
    \renewcommand{\arraystretch}{1.1}
    \begin{tabular*}{\columnwidth}{@{\extracolsep{\fill}}*{10}{c}@{}}
    \toprule
    \multicolumn{5}{c}{\textbf{Onboard}} &
    \multicolumn{5}{c}{\textbf{Offboard}} \\
    \cmidrule(lr){1-5}
    \cmidrule(lr){6-10}
     & Mean & Std. & $W_1 \downarrow$ &
     \shortstack{Norm.\\$W_1$ (\%)}
     & & Mean & Std. & $W_1 \downarrow$ &
     \shortstack{Norm.\\$W_1$ (\%)} \\
    \midrule
\textbf{CH} & \textbf{127.76} & \textbf{11.51} &
\textbf{3.02} & \textbf{2.42}
& \textbf{CH} & \textbf{415.50} & \textbf{34.33} &
\textbf{7.82} & \textbf{1.88} \\
\textbf{CM} & \textbf{127.72} & \textbf{11.51} &
\textbf{2.98} & \textbf{2.39}
& \textbf{CM} & \textbf{415.87} & \textbf{34.28} &
\textbf{7.61} & \textbf{1.83} \\
\textbf{SM} & \textbf{121.37} & \textbf{0.01} &
\textbf{10.28} & \textbf{8.24}
& \textbf{SM} & \textbf{226.05} & \textbf{18.99} &
\textbf{165.21} & \textbf{42.22} \\
\midrule
\textbf{GT} & 124.74 & 11.54 & -- & --
& \textbf{GT} & 416.86 & 35.89 & -- & -- \\
    \bottomrule
    \end{tabular*}
    \par\vspace{2pt}
    \begin{minipage}{\columnwidth}
    \footnotesize
    \textit{Note:} Mean, standard deviation, and $W_1$ are reported in milliseconds. Normalized $W_1$ is expressed as a percentage of the corresponding GT mean.
    \end{minipage}
    \end{table}

\subsection{Timing Fidelity Validation}
\label{sec:eval_timing}

We first validate C2Flight's timing fidelity using command age, defined as the end-to-end delay from sensor input to the availability of the corresponding control command. This metric tracks observation freshness during flight and captures how computing-system timing affects closed-loop behavior.
The validation compares the command-age distributions produced by CM and CH with those measured from the integrated physical GT pipelines during closed-loop runs.
Each configuration is evaluated in three independent runs covering more than 100 frames in total.

Table~\ref{tab:eval_timing} reports the mean, standard deviation, and empirical first Wasserstein distance ($W_1$) from GT.
We also normalize $W_1$ by the corresponding GT mean to express the distribution error relative to each system's average command age.
CM and CH closely reproduce the reference mean and variability, with normalized $W_1$ errors below 2.5\% onboard and 1.9\% offboard.
In contrast, SM's larger $W_1$ errors highlight the value of the parameterized dataflow graph and graph runtime scheduler in capturing timing effects from inter-node buffering, scheduling, and resource interactions.
These results confirm that C2Flight reproduces the distinct timing behaviors of both systems using independent hardware and component models, while retaining the same target graph across execution modes.

\subsection{Closed-Loop Task Prediction}
\label{sec:eval_closed_loop}

{\looseness=-1
We further evaluate navigation in an 8.0~m-long, 1.5~m-wide tunnel with four alternating obstacles to assess whether modeling computing-system timing improves closed-loop task-performance prediction.
The four obstacles are staggered along opposite walls, requiring successive left--right avoidance maneuvers, and their textures and dimensions match the real-world experimental setup of PULP-DroNet~\cite{lamberti2024pulpdronetIOTJ}.
The maximum forward speed is swept at a flight height of 0.3~m, with up to seven fixed-seed runs for each configuration--speed pair. Task success rate is defined as the fraction of runs in which the vehicle center reaches $x \geq 7.8$~m without collision before timeout.
\par}

{\looseness=-1
\textbf{Task prediction.}
Table~\ref{tab:eval_speed} reports the GT task-success rate and, for each evaluation mode, the prediction agreement rate, defined as the fraction of seed-matched runs whose predicted success or failure agrees with GT. 
For the onboard system at 1.0~m/s, GT completes 2/7 runs, compared with 0/7 and 1/7 for CM and CH, and 7/7 for both SC and SW.
The offboard results show the same separation at 1.00~m/s: CM and CH reproduce the reference failures, whereas SC and SW complete every tested run.
Both C2Flight modes therefore better predict the target systems' task limitations.
The transition is not exact: at 0.75~m/s offboard, CM is more pessimistic than GT, while CH gives a closer completion estimate.
\par}

\begin{table}[t]
\centering
\caption{Task-success rates across the speed sweep}
\label{tab:eval_speed}
\label{tab:eval_a2_speed}
\footnotesize
\setlength{\tabcolsep}{1pt}
\renewcommand{\arraystretch}{1.1}
\begin{minipage}[t]{0.49\columnwidth}
\vspace{0pt}
\centering
\begin{tabular*}{\linewidth}{@{\extracolsep{\fill}}cl|llll@{}}
\toprule
\multicolumn{6}{c}{\textbf{Onboard}} \\
\midrule
\shortstack{Speed (m/s)} & \textbf{GT} & \textbf{CH} & CM & SC & SW \\
\midrule
0.80 & 7/7 & \textbf{7/7} & 7/7 & 7/7 & 7/7 \\
0.90 & 7/7 & \textbf{7/7} & 6/7 & 7/7 & 7/7 \\
1.00 & 2/7 & \textbf{1/7} & 0/7 & 7/7 & 7/7 \\
1.10 & 1/7 & \textbf{0/7} & 0/7 & 5/7 & 7/7 \\
1.20 & 0/7 & \textbf{0/7} & 0/7 & 0/7 & 0/7 \\
\bottomrule
\end{tabular*}
\end{minipage}\hfill
\begin{minipage}[t]{0.49\columnwidth}
\vspace{0pt}
\centering
\begin{tabular*}{\linewidth}{@{\extracolsep{\fill}}cl|llll@{}}
\toprule
\multicolumn{6}{c}{\textbf{Offboard}} \\
\midrule
\shortstack{Speed (m/s)} & \textbf{GT} & \textbf{CH} & CM & SC & SW \\
\midrule
0.50 & 7/7 & \textbf{7/7} & 7/7 & 7/7 & 7/7 \\
0.75 & 3/7 & \textbf{2/7} & 0/7 & 7/7 & 7/7 \\
1.00 & 0/7 & \textbf{0/7} & 0/7 & 7/7 & 7/7 \\
1.25 & 0/7 & \textbf{0/7} & 0/7 & 0/7 & 0/7 \\
1.50 & 0/7 & \textbf{0/7} & 0/7 & 0/7 & 0/7 \\
\bottomrule
\end{tabular*}
\end{minipage}
\end{table}

\begin{table}[t]
\centering
\caption{Onboard trajectory errors at 1.0~m/s, averaged over seven paired runs relative to GT.}
\label{tab:eval_trajectory_errors}
\setlength{\tabcolsep}{5pt}
\footnotesize
\begin{tabular}{lrr}
\hline
Config. & Position MAE (cm) & Integrated error (m$\cdot$s) \\
\hline
CH & 3.040  & 0.326 \\
CM & 3.782  & 0.406 \\
SC & 12.456 & 1.335 \\
SW & 14.661 & 1.570 \\
\hline
\end{tabular}
\end{table}

\begin{figure}[t]
\centering
\includegraphics[
    width=\columnwidth,
    trim=4pt 4pt 4pt 4pt,
    clip
]{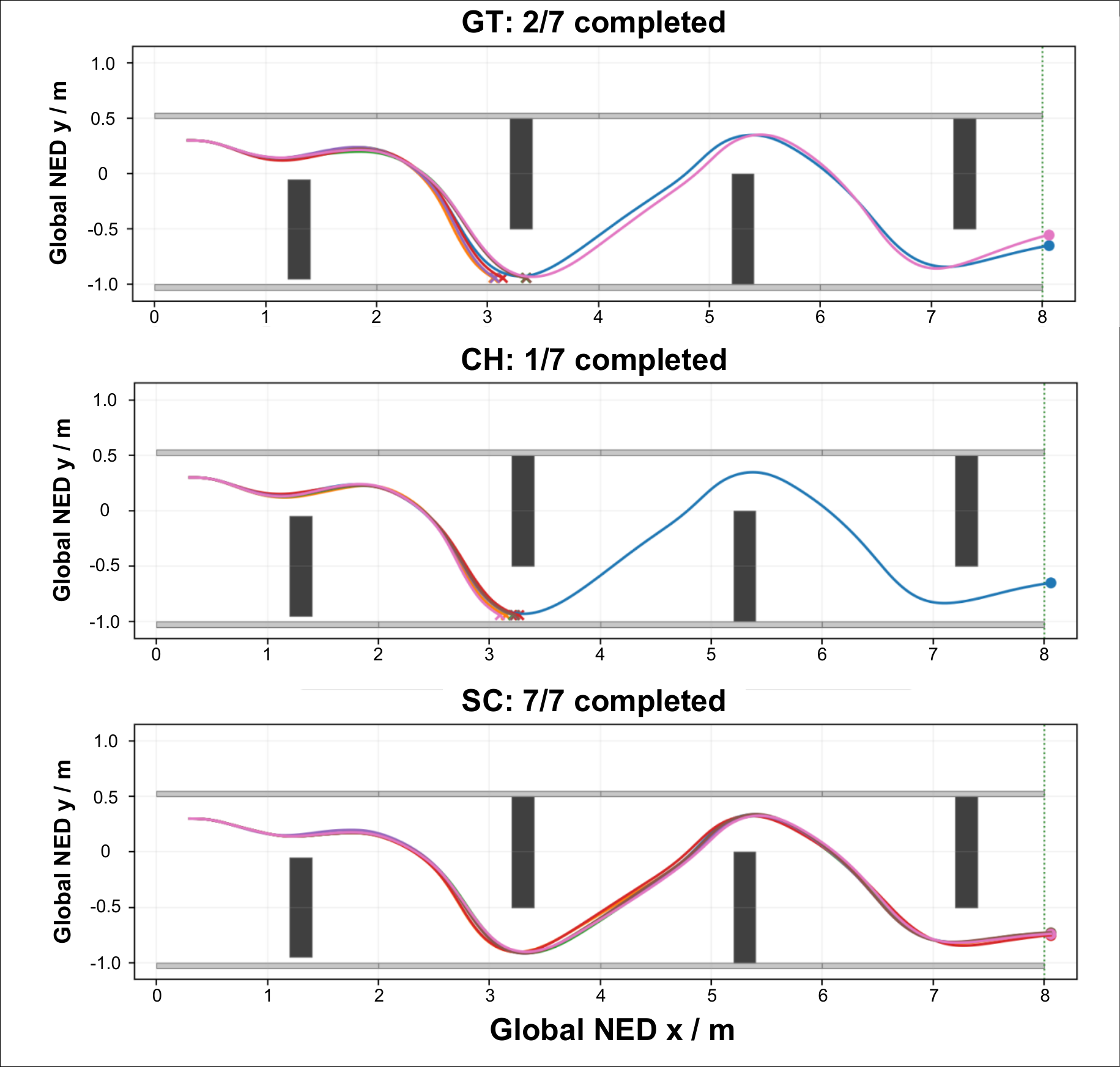}
\caption{Illustrative mean trajectories for the onboard system at 1.0~m/s.
Errors in Table~\ref{tab:eval_trajectory_errors} are computed from paired runs, not between mean curves.}
\label{fig:eval_trajectories}
\end{figure}

\textbf{Trajectory fidelity.}
Next, trajectory-level analysis is performed for the onboard system at 1.0~m/s, an operating point near the transition from reliable completion to frequent failure.
Table~\ref{tab:eval_trajectory_errors} reports errors over seven seed-matched runs, comparing positions at matched simulation timestamps.
The smaller SC/SW position MAE is 3.29 times the larger CM/CH error, with the same ordering in time-integrated error.
Both C2Flight modes thus better capture motion progression, beyond predicting the final task outcome. Figure~\ref{fig:eval_trajectories} provides an illustrative trajectory comparison at this operating point. Together, these results support predicting closed-loop task performance from independently modeled or executed components before full-system integration.


%% file: Sections/5_Case_Study.tex
\section{Case Study}

{\looseness=-1
C2Flight enables designers to compare candidate UAV computing systems based on closed-loop task performance before full-system integration.
We demonstrate this capability through two case studies: \emph{Onboard--Offboard Deployment Choice} and \emph{Onboard Accelerator Precision Selection}.
Together, these case studies provide design insights for computation placement and accelerator precision selection by revealing how system-level constraints shape timing, accuracy, and closed-loop performance.
Further analysis of the evaluation results reveals how system-level interactions shape closed-loop behavior.
\par}

\subsection{Onboard--Offboard Deployment Choice}
\label{sec:case_deployment}
Onboard computation uses embedded processors and local links such as UART to reduce communication overhead, but limited compute capacity can constrain inference speed and model complexity.
Offboard computation is usually less constrained by compute capacity but introduces additional communication delay and jitter over wireless links such as Wi-Fi.
In both deployments, UAV control performance depends strongly on the control update period and end-to-end latency, which computation, communication, and scheduling jointly determine.
These timing characteristics interact with algorithmic accuracy to shape closed-loop control behavior.

{\looseness=-1
To investigate this choice, we compare an onboard computing system and an offboard system using the CM modeling approach validated in Section~\ref{sec:evaluation}.
Both systems retain computation and communication parameters profiled from their physical implementations in the preceding evaluation. Image-acquisition request rates are swept from 2 to 20~Hz to compare closed-loop capability. Each rate is characterized by its effective control update period, end-to-end latency, and highest tested forward-speed cap with successful tunnel completion.
\par}

\begin{figure}[t]
    \centering
    \includegraphics[
        width=\columnwidth,
        trim=3pt 3pt 3pt 3pt,
        clip
    ]{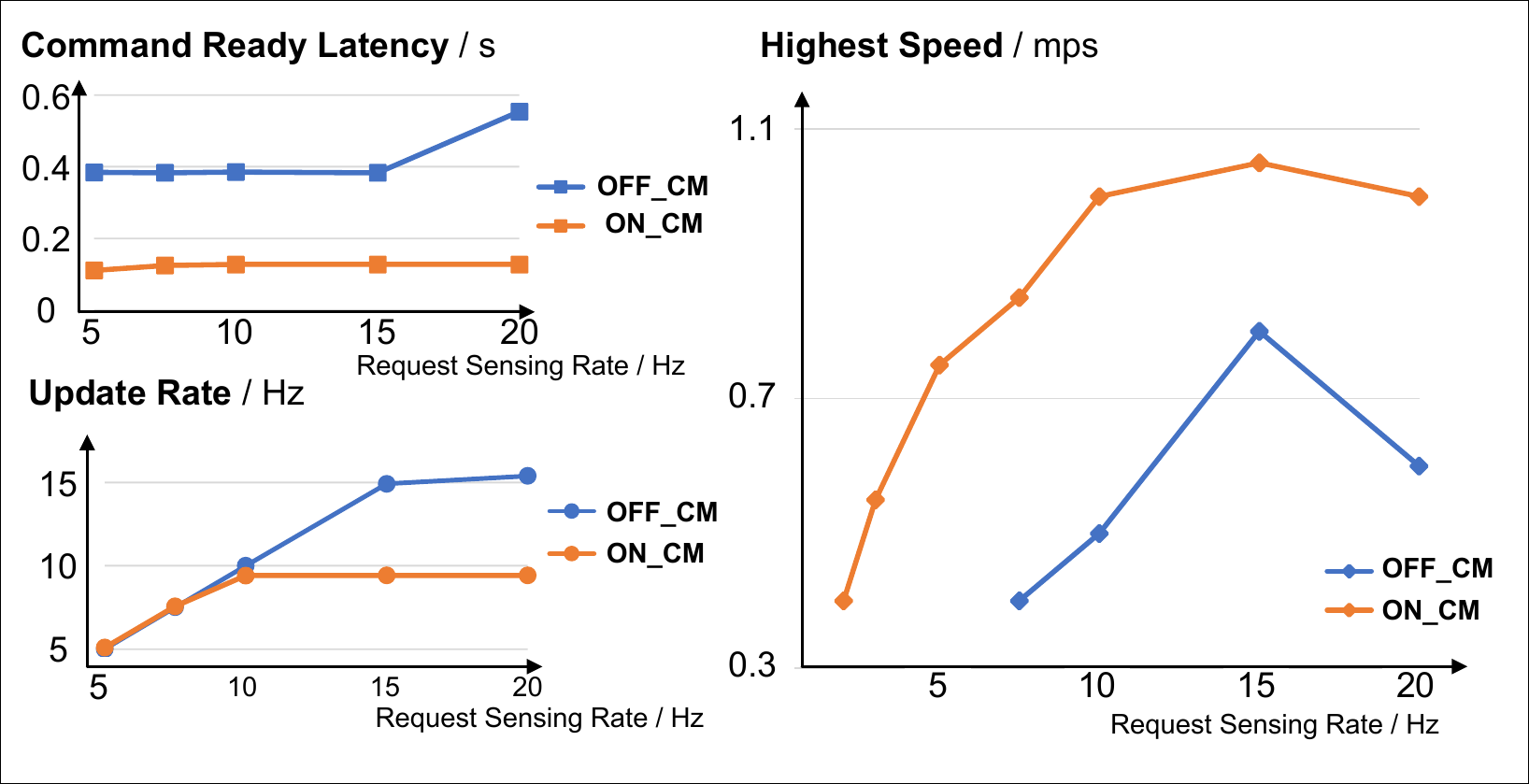}
    \caption{
        Onboard--offboard deployment exploration.
        Left: effective new-result rate and inverse command-ready latency,
        corresponding to the reciprocals of the control update period and
        end-to-end latency.
        Right: highest tested speed cap with at least one completed trial.
    }
    \label{fig:c2_deployment}
\end{figure}

{\looseness=-1
Figure~\ref{fig:c2_deployment} shows the interaction between command-ready latency and effective update rate.
As the sensing request rate increases, overlapping operations can introduce additional waiting and raise command-ready latency.
Meanwhile, the effective update rate increases and eventually plateaus at each system's throughput limit, with the offboard system supporting a higher rate.
Together, these timing characteristics determine a favorable operating range: excessive requests provide little additional update benefit while increasing latency, which can degrade control performance.
As shown in the lower panel of Fig.~\ref{fig:c2_deployment}, the highest observed speed curves identify approximately 15~Hz as a favorable common request-rate setting.
At this operating point, onboard computation achieves a higher successful speed cap despite its lower update rate, making it the preferred deployment under the evaluated constraints.
\par}

\subsection{Accelerator Bit-Width Selection}
\label{sec:c1}

Arithmetic bit width is another key consideration in the early-stage design of UAV processors.
Prior studies commonly adopt low-bit computation to reduce execution time and data-transfer costs~\cite{lamberti2024pulpdronetIOTJ}, but aggressive quantization can degrade algorithmic accuracy.
Wider formats preserve numerical fidelity but increase computation and memory-access costs, while data movement, memory bandwidth, and resource contention further shape command-ready latency. C2Flight evaluates how this coupled accuracy--latency trade-off affects closed-loop flight performance to guide bit-width selection.

To investigate this choice, CM models four Array--PPU RTL designs using INT4, INT8, FP16, and FP32 under a common approximate area budget.
Each candidate uses a format-specific functional model and latency distributions profiled from RTL simulation.
Following the architectural organization of NVDLA~\cite{nvdlaHardware}, we represent buffer accesses, DMA transfers, Array--PPU pipelining, and shared-resource contention in the system graph.
Each candidate is then integrated into the validated onboard system in place of GAP8 perception, while the DroNet architecture, UART communication, and STM32 navigation remain unchanged. A forward-speed-cap sweep with five trials per point identifies the highest tested cap achieving at least $4/5$ completions for each candidate.

\begin{table}[t]
    \centering
    \caption{
        Accelerator precision selection using CM
    }
    \label{tab:c1_precision}
    \small
    \setlength{\tabcolsep}{2pt}
    \begin{tabular}{c*{4}{c}}
        \toprule
        Speed cap
        & INT4 & INT8 & FP16 & FP32 \\
        \midrule
        $0.50$ m/s
        & 5/5 & 5/5 & 5/5 & 5/5 \\

        $0.75$ m/s
        & 5/5 & 5/5 & 4/5 & 2/5 \\

        $1.00$ m/s
        & 0/5 & \textbf{5/5} & 0/5 & 0/5 \\

        $1.25$ m/s
        & 0/5 & 0/5 & 0/5 & 0/5 \\
        \bottomrule
    \end{tabular}
    \par\vspace{2pt}
    \begin{minipage}{\columnwidth}
        \raggedright
        \footnotesize
        \textit{Note:} Entries denote completed/total trials. Boldface marks the selected precision at the highest tested speed cap meeting the $\geq 4/5$ completion criterion.
    \end{minipage}
\end{table}

Table~\ref{tab:c1_precision} shows that INT4 and INT8 outperform the wider formats at 0.75~m/s, but their performance diverges at higher speeds.
For this tunnel-navigation task, the observed pattern is consistent with a trade-off between accuracy loss from aggressive INT4 quantization and increased computation and data-movement overhead at larger bit widths.
INT8 is the best-performing candidate under the completion criterion, achieving $5/5$ completions at 1.0~m/s, where all other candidates fail.
C2Flight thus identifies INT8 as the preferred bit width under the evaluated area, memory-bandwidth, and resource constraints, prior to full accelerator RTL implementation and integration-level verification.

%% file: Sections/6_Conclusion.tex
\section{Conclusion}

This work presented C2Flight for closed-loop evaluation of heterogeneous UAV computing systems before full integration.
Its parameterized dataflow graph and unified node interface combine component models and independently accessible hardware while preserving system-level timing interactions.
Validation against integrated physical computing pipelines coupled to simulated flight showed close agreement in command timing and closed-loop behavior.
The case studies demonstrated how these evaluations guide compute placement and accelerator precision selection beyond isolated component metrics.
Future work will extend validation to physical flight and broader operating conditions. Component profiling and backend integration will be streamlined to support a wider range of computing platforms. Integrating C2Flight with automated design-space exploration would extend its role from evaluating candidate systems to jointly optimizing algorithms, hardware architectures, and compute placement based on closed-loop task performance.